\documentclass[11pt]{article}

\usepackage{dialogue}
\usepackage{booktabs}
\usepackage{ifxetex}
\ifxetex
    \usepackage{fontspec}
    \setromanfont{Times New Roman}
\else
  \usepackage{cmap}
  \usepackage[T2A]{fontenc}
  \usepackage[utf8]{inputenc}
  \usepackage{times}
  \usepackage{latexsym}
  \usepackage{substitutefont}
  \substitutefont{T2A}{\familydefault}{cmr}
\fi

\usepackage[russian,british]{babel}
\usepackage{url}
\usepackage{pgf}
\usepackage{multirow}
\usepackage{comment}
\usepackage{tabularx}
\usepackage[justification=centering]{caption}

\usepackage{covington} 

\usepackage{hyperref}

\DeclareRobustCommand{\cyrins}[1]{%
  \begingroup\fontfamily{erewhon-TLF}%
  \foreignlanguage{russian}{#1}%
  \endgroup
}

\dialogfinalcopy 

\title{RuArg-2022: Argument Mining Evaluation}

\author{\textbf{Evgeny Kotelnikov}$^1$, \textbf{Natalia Loukachevitch}$^2$, \textbf{Irina Nikishina}$^3$, \textbf{and} \\
\textbf{Alexander Panchenko}$^3$,    \\
$^1$Vyatka State University, $^2$Lomonosov Moscow State University, \\ 
$^3$Skolkovo Institute of Science and Technology \\
\href{mailto:kotelnikov.ev@gmail.com}{kotelnikov.ev@gmail.com}, \href{mailto:louk_nat@mail.ru}{louk\_nat@mail.ru}, \\
\href{mailto:a.panchenko@skoltech.ru}{\{irina.nikishina, a.panchenko\}@skoltech.ru} }

\date{}

\begin{document}
\maketitle
\begin{abstract}

Argumentation analysis is a field of computational linguistics that studies methods for extracting arguments from texts and the relationships between them, as well as building argumentation structure of texts. This paper is a report of the organizers on the first competition of argumentation analysis systems dealing with Russian language texts within the framework of the Dialogue conference. During the competition, the participants were offered two tasks: stance detection and argument classification. A corpus containing 9,550 sentences (comments on social media posts) on three topics related to the COVID-19 pandemic (vaccination, quarantine, and wearing masks) was prepared, annotated, and used for training and testing. The system that won the first place in both tasks used the NLI (Natural Language Inference) variant of the BERT architecture, automatic translation into English to apply a specialized BERT model, retrained on Twitter posts discussing COVID-19, as well as additional masking of target entities. This system showed the following results: for the stance detection task an F1-score of 0.6968, for the argument classification task an F1-score of 0.7404. We hope that the prepared dataset and baselines will help to foster further research on argument mining for the Russian language. 
  
  
  \textbf{DOI:} 10.28995/2075-7182-2022-21-333-348
\end{abstract}

\selectlanguage{russian}
\begin{center}
  \russiantitle{RuArg-2022: соревнование по анализу аргументации}

 

\begin{tabular}{c}
\textbf{Евгений Котельников}$^1$, \textbf{Наталья Лукашевич}$^{2}$,
\textbf{Ирина Никишина}$^3$, \textbf{и} \\ \textbf{Александр Панченко}$^1$ \\
$^1$Вятский государственный университет, \\
$^2$Московский государственный университет им. М.В.Ломоносова, \\
$^3$Сколковский институт науки и технологий \\
\href{mailto:kotelnikov.ev@gmail.com}{kotelnikov.ev@gmail.com}, \href{mailto:louk_nat@mail.ru}{louk\_nat@mail.ru}, \\
\href{mailto:a.panchenko@skoltech.ru}{\{irina.nikishina, a.panchenko\}@skoltech.ru}
\end{tabular}
  
\end{center}

\begin{abstract}
Анализ аргументации – это область компьютерной лингвистики, в которой исследуются методы извлечения из текстов аргументов и связей между ними, а также построения аргументационной структуры. Настоящая статья представляет собой отчет организаторов о первом соревновании русскоязычных систем анализа аргументации в рамках конференции «Диалог». В ходе соревнования участникам были предложены две задачи: определение позиции автора по заданной теме и классификация доводов. Для обучения и тестирования систем был подготовлен и размечен корпус, содержащий 9,550 предложений (комментариев к постам в социальных медиа) по трем тематикам, связанным с пандемией COVID-19:  вакцинация, карантин и ношение масок. Система, занявшая первое место по обеим задачам, использовала NLI вариант (Natural Language Inference -- вывод по тексту) применения архитектуры BERT, автоматический перевод на английский язык для использования специализированной модель BERT, дообученной на постах Твиттера, обсуждающих ковид, а также дополнительное маскирование целевых сущностей. Эта система показала следующие результаты: для задачи определения позиции F1-score=0.6968, для задачи классификации доводов F1-score=0.7404. Мы надеемся, что подготовленные наборы данных и методы помогут стимулировать дальнейшие исследования по анализу аргументации для русского языка.
  
  \textbf{Ключевые слова:} анализ аргументации, определение позиции автора текста, классификация доводов, COVID-19
\end{abstract}
\selectlanguage{british}

\section{Introduction}
\label{introduction}

%
%

People have been constantly arguing at all social levels and the
\textit{Argumentation Theory} was developed to study and control the process of coming to a conclusion from premises through logical reasoning. According to this theory, an argument must include a \textit{claim} containing a \textit{stance} towards some topic or object, and at least one \textit{premise} (``favor'' or ``against'') of this stance. Often a ``premise'' is called an ``argument'' when it is clear from the context which claim it is being referred to.

With the development of intellectual systems and neural networks, arguments can now be both produced and studied automatically. Therefore, the \textit{Computational Argumentation} task arose to address the problem of computational analysis and synthesis of natural language argumentation. In this paper we focus on its branch --- \textit{Argument Mining} (or \textit{Argumentation Mining}) --- which explores methods for extracting arguments and their relationships from texts, as well as constructing an argumentative structure.


There is a large number of works devoted to this field which are thoroughly reviewed by  
\cite{Stede-Schneider2018,Lawrence-Reed:2020,Stede2020,Vecchi-etal:2020,Schaefer-Stede2021}.
Special attention has been paid to the stance detection as a sub-task of Argument Mining \cite{Kucuk-Can2020,ALDayel-Magdy2021,Kucuk-Can2021} where the authors describe the proposed approaches so far, descriptions of the relevant datasets and tools, and some other related 
issues.

The growing interest in the task is justified by the application of the Argument Mining algorithms for argument search, fact checking, automated decision making, argument summarization, writing support and intelligent person assistants. For instance, Args.me\footnote{\href{https://www.args.me/index.html}{https://www.args.me/index.html}},  ArgumenText\footnote{\href{https://www.informatik.tu-darmstadt.de/ukp/research_ukp/ukp_research_projects/ukp_project_argumentext/index.en.jsp}{https://www.informatik.tu-darmstadt.de/ukp/research\_ukp/ukp\_research\_projects/ukp\_project\_argumentext/index.en.jsp}} and CAM (comparative argumentative machine)\footnote{\href{http://ltdemos.informatik.uni-hamburg.de/cam/}{http://ltdemos.informatik.uni-hamburg.de/cam/}}\cite{cam} are well known systems widely used for searching arguments.

The main research forum for the task is the Argument Mining workshop series.
Since 2014, eight workshops on the analysis of arguments have already been held\footnote{\href{https://2021.argmining.org}{https://2021.argmining.org}} addressing burning issues like multi-task learning \cite{tran-litman-2021-multi,putra-etal-2021-multi} and Argumentation Mining in different areas (science \cite{lauscher-etal-2018-arguminsci,fergadis-etal-2021-argumentation}, news articles \cite{bauwelinck-lefever-2020-annotating} and cross-lingual research \cite{rocha-etal-2018-cross}). Moreover, there are several shared tasks on the topic adjacent to the Argument Mining: Shared Task on Argumentation Mining in Newspaper Editorials \cite{Kiesel-etal2015}, SemEval-2016 (Stance Detection) \cite{Mohammad-etal2016}, Touché (Argument Retrieval) in 2020 \cite{Bondarenko-etal:2020} and 2021 \cite{Bondarenko2021}.

In this paper, we present RuArg-2022 — the first shared task on Argument Mining for the Russian language. It consists of two sub-tasks: stance detection and premise classification. The first task aims to determine the point of view (stance) of the text’s author in relation to the given claim. The second task is devoted to  classification of texts according to premises (``for'' or ``against'') to a given claim.

To highlight the differences between the two tasks, consider the following example: \textit{\cyrins{Я против масок, но приходится их носить: мне проще так, чем с кем-то что-то обсуждать и кому-то что-то доказывать}} (\textit{I am against masks, but I have to wear them: it’s easier for me than to discuss something with someone and prove something to someone}). In this sentence there is an explicit stance against masks but it gives a premise for wearing masks.

The contribution of the current paper is three-fold. First, we prepare a gold standard dataset for stance detection and premise classification. 
Second, we develop and release a baseline for the argument mining tasks that uses a multi-task multi-label BERT architecture. Third, we compare and analyse the results of the participants of the shared task and propose steps for further improvement of both sub-tasks. All the materials and data could be found on GitHub\footnote{\href{https://github.com/dialogue-evaluation/RuArg}{https://github.com/dialogue-evaluation/RuArg}} and CodaLab\footnote{\href{https://codalab.lisn.upsaclay.fr/competitions/786}{https://codalab.lisn.upsaclay.fr/competitions/786}} competition pages.

Thus, our work is the first, to the best of our knowledge, dealing with argument mining task for the Russian language. While our setup is a simple text categorization task, we argue that it may be an important building block of larger argument mining pipelines  featuring  retrieval of arguments from large text collections~\cite{Bondarenko-etal:2020}.

\section{Previous Work}
\label{previous}

The Argument Mining task \cite{argumentation} involves the automatic identification of argumentative structures in free text. According to \cite{argmining-2020-argument}, ``researchers have investigated argument mining on various registers including legal texts, scientific papers, product reviews, news editorials, Wikipedia articles, persuasive essays, political debates, tweets, and online discussions''. 
A detailed overview of all argument mining related papers is out of the scope of the current work. We refer the reader to the recent surveys on this topic: \cite{Lawrence-Reed:2020} and \cite{Schaefer-Stede2021}.

The topic of COVID-19 is nowadays popular not only in the biomedical field, but also in social science and, especially, NLP research \cite{nlp-covid19-2020-nlp-covid}. There already exist several datasets on COVID-19 for stance detection \cite{wuhrl-klinger-2021-claim}, argument mining and fact extraction/verification. For instance, in
\cite{beck-etal-2021-label-suggestions} the authors collect a dataset from German Twitter on people's attitude towards the government measures. First, they identify relevant tweets for governmental measures and if relevant, detect what stance is expressed. \cite{argmine_covid} create a linked data version of the CORD-19 data set and enriched it via entity linking and argument mining.

Another dataset collected lately \cite{reddy2021newsclaims} comprises the following topics related to COVID-19: origin of the virus, transmission of the virus, cure for the virus and protection from the virus. The authors present a pipeline for detecting claim boundaries and detecting stance. Unlike most stance detection datasets \cite{hanselowski-etal-2019-richly,allaway-mckeown-2020-zero} it involves identifying the claimer’s stance within a claim sentence and not the stance for target–context pairs. \cite{licovid} follows \cite{reddy2021newsclaims} and identify the stance from the perspective of each claimer, namely whether the claimer affirms or refutes a claim. They finetune a Bart-large model \cite{lewis-etal-2020-bart} to automatically identify the stance. 
One more dataset related to the COVID-19 pandemic is collected by seven science teachers through three scenarios \cite{atabey2021science}. This dataset contains not only stances about vaccination, curfew and distance education, but also arguments and supporting reasons that might construct an argument mining dataset.

Most of the above mentioned publications on COVID-19 stance detection refer to the FEVER-like dataset COVID-Fact  \cite{Saakyan2021COVIDFactFE} of 4,086 claims concerning the COVID-19 pandemic. It could be also applied for the argument mining needs. Another dataset for COVID-19 fact checking is presented in \cite{liu-etal-2020-adapting-open} which also could be reformatted for the argument mining tasks.

As regards argumentation mining for the Russian language, there are not so many studies and datasets on the topic. \cite{Fishcheva2019} translated into Russian and researched the English language Argumentative Microtext Corpus (ArgMicro) \cite{Peldszus2015,Skeppstedt2018}. In \cite{Fishcheva-etal2021} this corpus was expanded with machine translation of the Persuasive Essays Corpus (PersEssays) \cite{Stab2014}. XGBoost and BERT were applied to classify “for”/“against” premises.

Salomatina et al. \cite{Salomatina2021} propose an approach to the partial extraction of the argumentative structure of a text by using patterns of
argumentation indicators. They also try to recognize the relations between extracted arguments. \cite{Ilina2021} develop a web resource for analysis of argumentation in popular science discourse. The annotation model is based on the ontology of argumentation and D. Walton's argumentation schemes \cite{Walton-etal2008}. A scenario of argument annotation of texts allows constructing an argumentative graph based on the typical reasoning schemes.

To the best of our knowledge, there are no manually labelled publicly available datasets in Russian. In this work we present such dataset for the first time.

\section{Dataset}
\label{datasets}


The dataset is based on VKontakte users' comments discussing COVID-2019 news texts \cite{chkhartishvili2021covid}. We choose the COVID-19 pandemic (and anti-epidemic measures in general) as the topic of the dataset because we assume that the analysis of arguments on social measures against COVID-19 is still relevant for the modern society and especially for the understanding of the current public sentiment. 
From the gathered comment collection, sentences discussing masks, vaccines and quarantine were extracted using keywords  \cite{nugamanov2021extracting}. 

The annotation process included two stages: labelling by stance and labelling by premises. At both stages sentences were labelled in relation to the following claims:
\begin{enumerate}
    \item “Vaccination is beneficial for society.”
    \item “The introduction and observance of quarantine is beneficial for society.”
    \item “Wearing masks is beneficial for society.”
\end{enumerate}

In the following subsections we describe the annotation process of the dataset for both sub-tasks: Stance Detection \ref{stance} and Premise Classification \ref{premise}.

\begin{table}[h]
\centering
\begin{tabular}{|c|c|c|}
\toprule
\bf Stance & \bf Premise & \bf Numerical label\\
\midrule
 for & for & 2\\
other (neutral/contradictory/unclear) & no argument & 1\\
against & against& 0 \\
irrelevant & irrelevant & -1 \\
\bottomrule
\end{tabular}
\caption{System of categories used to label the dataset.}
\label{tab:coding} 
\end{table}

\subsection{Stance Annotation}
\label{stance}

The current dataset has been already annotated in \cite{nugamanov2021extracting}. In the current work the dataset was additionally checked and synchronized with premise annotation from the second step.

At the first stage of stance annotation each sentence was labelled by several experts (three on average).  An annotator should indicate the stance it expresses towards each of the above-mentioned aspects (or indicate that the sentence is not relevant to the aspect). The annotators' group included professional  linguists and psychologists.
We consider four stance labels, namely:
\begin{itemize}
    \item {\bf for:} positive stance, which means that the speaker expresses his support for the topic;
    \item {\bf against:} negative stance --- the topic of discussion is not endorsed by the speaker;
    \item {\bf other:} neutral stance (this label is used for factual sentences without any visible attitudes from the author); contradictory stance (for such a label, evident positive and negative attitudes should be seen in a message); unclear stance (the presence of a stance is seen, but the context of sentence does not give possibility to determine it);
    \item {\bf irrelevant:} text does not contain stance on the topic.
\end{itemize}

The coding scheme for the stance annotation is presented in Table \ref{tab:coding}.

A sentence is considered to be relevant to an aspect, if at least two annotators considered it relevant. Sentences collected using keywords also can be irrelevant, for example a sentence mentioning Elon Musk (``Mask'' in Russian spelling) is not relevant to the mask aspect. 

\subsection{Premise Annotation}
\label{premise}

At the second stage of annotation, the dataset was also annotated by premises for all three claims. The following four classes (labels) were used:

\begin{itemize}
    \item {\bf for:} the stance is supported with argument in favor of the topic;
    \item {\bf against:} the argument explains the author's negative outlook on the topic;
    \item {\bf no argument:} no explanation is given for supporting/critisism of the topic;
    \item {\bf irrelevant:} text does not contain stance and, consequently, premise on the topic.
\end{itemize}

The annotated sentences from the previous step were divided into three subsets: training, validation, and test (see Subsection~\ref{dataset_stats}). The labelling of each sentence by premises from the training and validation datasets was carried out by three annotators; the test sentences were labelled by four annotators. The final labels for training and validation datasets were assigned with the agreement of at least two annotators, for test dataset – with the agreement of at least three annotators.

A sentence was considered as a premise if the annotator could use it to convince an opponent about the given claim, such as ``\textit{Masks help prevent the spread of disease.}'' Detailed instructions for annotators are available in the competition repository\footnote{\href{https://github.com/dialogue-evaluation/RuArg/tree/main/annotation}{https://github.com/dialogue-evaluation/RuArg/tree/main/annotation}.}.

The task of premise annotation should be separated from stance detection and sentiment analysis tasks. For example, the following statement does not contain a premise in relation to masks, although there is an author’s stance ``for'': \textit{It is high time to involve the city of ``brides'' in the production of protective masks}. It is also necessary to distinguish between sentiment polarity (positive and/or negative) and argumentation. In the following sentence there is a negative polarity towards quarantine, a positive polarity towards Trump, but no rational premises ``for'' or ``against'' quarantine are given: \textit{And the fact that Trump did not introduce a suffocating quarantine is well done!}

The difference between the two tasks is illustrated in Table \ref{examples}. For example, the first sentence possesses both stance and premise: the speaker expresses his negative attitude towards the vaccine by the reason of its short-term effectiveness. The second sentence, on the contrary, definitely supports vaccination without giving any specific arguments to support his/her opinion.

\begin{table}[ht]
\centering
\footnotesize
\resizebox{\textwidth}{!}{  
\begin{tabular}{|l|c|c|c|c|c|c|}
\toprule
\multicolumn{1}{|c|}{Text}                                                                                                                                                                                                                                                               & \multicolumn{2}{c|}{Masks}                                 & \multicolumn{2}{c|}{Quarantine}                            & \multicolumn{2}{c|}{Vaccines}                           \\ \cmidrule{2-7}
\multicolumn{1}{|c|}{}                                                                                                                                                                                                                                                                                    & \multicolumn{1}{c|}{Stance} & \multicolumn{1}{c|}{Premise} & \multicolumn{1}{c|}{Stance} & \multicolumn{1}{c|}{Premise} & \multicolumn{1}{c|}{Stance} & \multicolumn{1}{c|}{Premise} \\ \midrule
\begin{tabular}[c]{@{}l@{}}\cyrins{И какой смысл в вакцине если антитела} \\ \cyrins{только 3 месяца?}\\ \textit{(And what's the point of a vaccine if the antibodies} \\ \textit{work only for 3 months?)}\end{tabular}                                                                                                                          & —                 & —                   & —                 & —                   & against                    & against                      \\ \midrule
\begin{tabular}[c]{@{}l@{}}\cyrins{Должна быть вакцина которую, будут} \\ \cyrins{прививать с детства!!!}\\ \textit{(There must be a vaccine that will be vaccinated} \\ \textit{from childhood!!!)}\end{tabular}                                                                                                                                 & —                 & —                   & —                 & —                   & for                        & no argument             \\ 
\midrule
   \begin{tabular}[c]{@{}l@{}}\cyrins{Вот только там на момент, когда была 1000 }\\ \cyrins{выявленных, уже неделю карантин действовал.}\\ \textit{(At the time when there were 1000} \\ \textit{identified, quarantine had been in effect for a week.)}\end{tabular}                          & —                 & —                   & other                        & against                          & —                 & —                   \\ \midrule 
\begin{tabular}[c]{@{}l@{}}\cyrins{Развитие ситуации: если соблюдать карантин}\\ \cyrins{месяц, то вирус будет остановлен.}\\ \textit{(The development of the situation: if the quarantine} \\ \textit{is observed for a month, the virus will be stopped.)}\end{tabular}                                                                        & —                 & —                   & for                        & for                          & —                 & —                   \\ \midrule
\begin{tabular}[c]{@{}l@{}}\cyrins{Вопрос к властям :почему из гос резерва} \\ \cyrins{не получили люди масок когда их не хватало} \\ \cyrins{или и резерва уже нет}\\ \textit{(Question to the authorities : why didn't people} \\  \textit{get masks from the state reserve when}\\ \textit{there were not enough of them or there is} \\  \textit{no reserve anymore)}\end{tabular} & for                        & no argument                  & —                 & —                   & —                 & —                   \\ \midrule
\begin{tabular}[c]{@{}l@{}}\cyrins{Любители масок не ужели вы думаете что эта} \\ \cyrins{косметическая тряпочка поможет от вируса?!}\\ \textit{(Mask lovers don't you really think that this cosmetic} \\ \textit{rag will help against the virus?!)}\end{tabular}                                                                               & against                    & no argument                  & —                 & —                   & —                 & —                   \\
\bottomrule
\end{tabular}}
\caption{Examples for each topic — masks, vaccines, and quarantine (we keep the original spelling and punctuation). Note that for each topic, annotation of stances and premises was performed. Refer to Table~\ref{tab:coding} for the classification schema used to label the data. ``Irrelevant'' class is denoted as ``—''.}
\label{examples}
\end{table}

\subsection{Dataset Verification}

After completion of stance and premise annotation procedures, we verified the labels of the dataset. To this end, we looked through the contingency tables of stances and premises, and checked the following issues:
\begin{enumerate}
    \item the sentence with an irrelevant label for one of the sub-tasks cannot be relevant for another sub-task;
    \item the sentence with contradictory stance and premise (e.g., positive stance but premise “against”) should be examined more carefully.
\end{enumerate}

As a result, annotations for 289 sentences (3.0\% from the whole dataset containing such issues) were revised and improved. Generally, if a sentence contains both a stance and a premise, then their polarity coincides (both “for” or both “against”). However, in 12 sentences the polarity is opposite. This is due to the fact that the sentence simultaneously contains the author’s point of view and indicates the opponents’ premises, for example: “\textit{This is exactly why everyone should wear masks, but the main channels broadcast that masks are not needed and useless for healthy people.}”

\subsection{Dataset Statistics}
\label{dataset_stats}

Each sentence has 6 labels: for each of the two sub-tasks (stance detection and premise classification) there is a label for each of the three aspects (masks, vaccines and quarantine).



The inter-annotator agreement was calculated by Krippendorff’s alpha and it turned out quite high – $0.84$. 
Dataset statistics are presented in Table~\ref{tab:statistics}.  As one may observe the dataset is skewed (imbalanced). There are various schemes in the literature to perform evaluation of this kind of data \cite{rosenberg}. We resort to a scheme by simply excluding the largest ``irrelevant'' class  as it is done in Sentiment Analysis. For instance, at the SemEval-2016 Task 14 \cite{nakov-etal-2016-semeval} the organizers exclude the ``NEUTRAL'' class from the evaluation as the largest one.

The distribution of labels by class is shown in Figure \ref{fig:results} 

\begin{table}[h!]
\begin{center}
\begin{tabular}{|l|l|l|l|l|l|l|l|l|}
\toprule
\multirow{2}{*}{Dataset} & \multirow{2}{*}{Total}
 & \multicolumn{3}{c|}{Stance} & \multicolumn{3}{c|}{Premise} & \multicolumn{1}{c|}{\multirow{2}{*}{Irrelevant}} \\ \cmidrule{3-8}
\multicolumn{1}{|c|}{} & \multicolumn{1}{c|}{} & \multicolumn{1}{c|}{For} & \multicolumn{1}{c|}{Other} & \multicolumn{1}{c|}{Against} & \multicolumn{1}{c|}{For} & \multicolumn{1}{c|}{No argument} & \multicolumn{1}{c|}{Against} & \multicolumn{1}{c|}{} \\ \midrule
\multicolumn{9}{|c|}{\textit{Masks}} \\
\midrule
train & 6,717 & \multicolumn{1}{r|}{704} & \multicolumn{1}{r|}{1,832} & 594 & \multicolumn{1}{r|}{339} & \multicolumn{1}{r|}{2,451} & 340 & 3,587 \\ 
val & 1,431 & \multicolumn{1}{r|}{148} & \multicolumn{1}{r|}{388} & 126 & \multicolumn{1}{r|}{62} & \multicolumn{1}{r|}{542} & 58 & 769 \\ 
test & 1,402 & \multicolumn{1}{r|}{147} & \multicolumn{1}{r|}{401} & 123 & \multicolumn{1}{r|}{63} & \multicolumn{1}{r|}{523} & 85 & 731 \\ 
all & 9,550 & \multicolumn{1}{r|}{999} & \multicolumn{1}{r|}{2,621} & 843 & \multicolumn{1}{r|}{464} & \multicolumn{1}{r|}{3,516} & 483 & 5,087 \\ \midrule
\multicolumn{9}{|c|}{\textit{Quarantine}} \\
\midrule
train & 6,717 & \multicolumn{1}{r|}{587} & \multicolumn{1}{r|}{1 341} & 172 & \multicolumn{1}{r|}{217} & \multicolumn{1}{r|}{1,756} & 127 & 4,617 \\ 
val & 1,431 & \multicolumn{1}{r|}{125} & \multicolumn{1}{r|}{290} & 39 & \multicolumn{1}{r|}{46} & \multicolumn{1}{r|}{369} & 39 & 977 \\
test & 1,402 & \multicolumn{1}{r|}{116} & \multicolumn{1}{r|}{274} & 40 & \multicolumn{1}{r|}{50} & \multicolumn{1}{r|}{358} & 22 & 972 \\ 
all & 9,550 & \multicolumn{1}{r|}{828} & \multicolumn{1}{r|}{1,905} & 251 & \multicolumn{1}{r|}{313} & \multicolumn{1}{r|}{2,483} & 188 & 6,566 \\ \midrule
\multicolumn{9}{|c|}{\textit{Vaccines}} \\
\midrule
train & 6,717 & \multicolumn{1}{r|}{374} & \multicolumn{1}{r|}{866} & 418 & \multicolumn{1}{r|}{149} & \multicolumn{1}{r|}{1,238} & 271 & 5,059 \\ 
val & 1,431 & \multicolumn{1}{r|}{78} & \multicolumn{1}{r|}{183} & 92 & \multicolumn{1}{r|}{24} & \multicolumn{1}{r|}{282} & 47 & 1,078 \\ 
test & 1,402 & \multicolumn{1}{r|}{75} & \multicolumn{1}{r|}{181} & 81 & \multicolumn{1}{r|}{21} & \multicolumn{1}{r|}{262} & 54 & 1,065 \\ 
all & 9,550 & \multicolumn{1}{r|}{527} & \multicolumn{1}{r|}{1,230} & 591 & \multicolumn{1}{r|}{194} & \multicolumn{1}{r|}{1,782} & 372 & 7,202 \\ 
\bottomrule
\end{tabular}
\end{center}
\caption{Statistics of the constructed dataset used in RuArg-2022 shared task.}
\label{tab:statistics} 
\end{table}

\begin{figure}[t]
  \centering
  \includegraphics[width=0.9\linewidth]{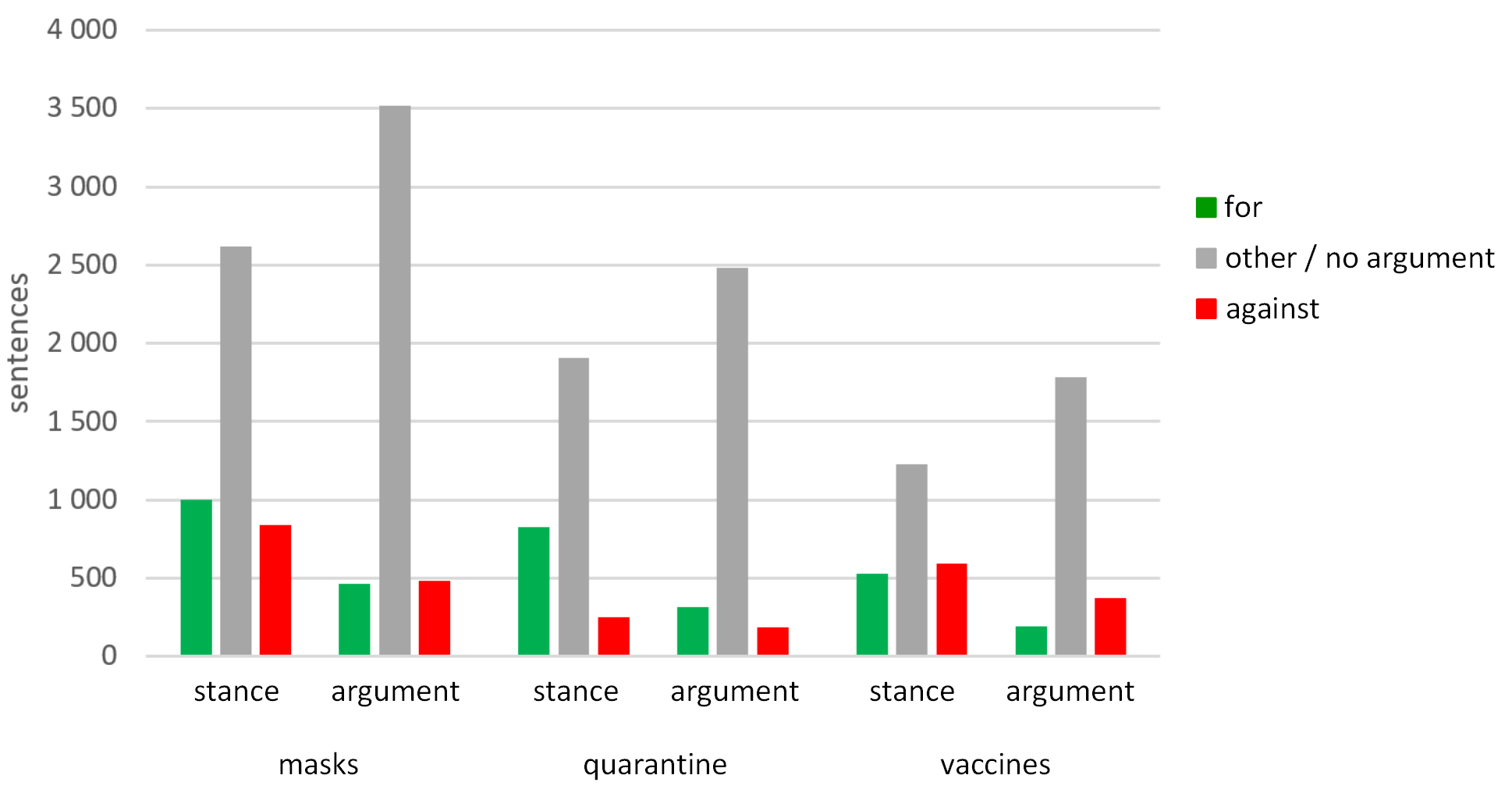}
  \caption{Distribution of the labels.}
  \label{figDatasets}
\end{figure}

\section{Evaluation}

The main performance metric in each of the two sub-tasks are F1$_{stance}$ and F1$_{premise}$ scores, which are calculated according to the following formula:

\begin{equation}
F1 = \frac{1}{n}\sum_{c \in C}F1_{rel_{c}}, 
\end{equation}

where $C=\{$``$masks$'', ``$vaccines$'', ``$quarantine$''$\}$, $n$ is the size of $C$ and F1$_{rel}$-score is macro F1-score averaged over first three relevance classes (the class ``irrelevant'' is excluded). Namely, the following procedure is used:

\begin{enumerate}
    \item F1-scores are averaged over three out of four classes (the ``irrelevant'' class is excluded) – macro F1$_{rel}$-score is obtained for a given claim;
    \item macro F1$_{rel}$-scores for all three claims are averaged – we get macro F1-score relative to the task (stance detection or premise classification);
    \item For each of the three claims, F1-score is calculated for each class (label) separately.
\end{enumerate}

As a result, two main macro F1$_{rel}$-scores are calculated – one for each sub-task.  Participants’ systems are ranked by these metrics (two separate lists). The F1$_{rel}$-score for claims and F1-score for individual classes (labels) will be also discussion in Section \ref{sec:results}.

\section{Baseline}

We implement a simple baseline that finetunes the pre-trained ruBERT model \cite{devlin-etal-2019-bert,Kuratov2019AdaptationOD} on the provided dataset. We chose ``DeepPavlov/rubert-base-cased'' model from Hugging Face \footnote{\href{https://huggingface.co/DeepPavlov/rubert-base-cased}{https://huggingface.co/DeepPavlov/rubert-base-cased}}. We experiment with training a single model that predicts all the required labels. However, it did not performed well, so we finetune three pre-trained BERT models separately for three topics: ``masks'', ``vaccines'', and ``quarantine''. Each model comprises the following layers:

\begin{enumerate}
    \item the pre-trained BERT layer with the unfrozen weights;
    \item a dense layer for stance detection;
    \item a dense layer for argument classification.
\end{enumerate}

Then we applied categorical cross-entropy loss to train on both stance and argument labels simultaneously. The results are presented in Section \ref{sec:results}.

\section{Participating Systems}
\label{participants}

RuArg-2022 shared task attracted 16 participants, 13 of them participated in the final phase. We provide descriptions of the top 7 solutions which outperformed the baseline for at least one sub-task. We denote each team either with its team name (if any) or with their CodaLab user names. In cases of multiple submissions from one team, we report only the best result. The scores of the teams are shown in Table \ref{tab:stance}.

\paragraph{camalibi (msu)} First, this team used RuBERT-classifier\footnote{\href{https://huggingface.co/DeepPavlov/rubert-base-cased-conversational}{https://huggingface.co/DeepPavlov/rubert-base-cased-conversational}} to determine the relevance of the texts using NLI-method: to form an input example, a second sentence with the aspect (``masks'', ``quarantine'', or ``vaccination'') was added to each original sentence from the dataset. The output $1$ was for the ``Relevant'' result and $0$ for ``Irrelevant''.

For the stance classification task the texts were pre-processed and then translated into English using pretrained seq2seq-model\footnote{\href{https://huggingface.co/Helsinki-NLP/opus-mt-ru-en}{https://huggingface.co/Helsinki-NLP/opus-mt-ru-en}}. Then, each text was processed according to the rule: $keyword \rightarrow @ * ASPECT * keyword \  @$, where $ASPECT$ is the aspect for which a given text is relevant and $keyword$ is the word from a list of words corresponding to each aspect.

Then for both RuArg sub-tasks the domain-specific BERT-classifier\footnote{\href{https://huggingface.co/digitalepidemiologylab/covid-twitter-bert-v2}{https://huggingface.co/digitalepidemiologylab/covid-twitter-bert-v2}} was trained using NLI-method: for each text and each aspect for which a given text is relevant, six input examples were constructed (three for each stance label and three for each premise label). Final input examples looked like: 
\begin{itemize}
    \item ``Vacation would only give rise to the spread of the virus, and it was not the weekend that had to be declared but the @ * quarantine * quarantine @.'', \textbf{``Against quarantine''},
    \item ``Vacation would only give rise to the spread of the virus, and it was not the weekend that had to be declared but the @ * quarantine * quarantine @.'', \textbf{``None-stance quarantine''},
    \item ``Vacation would only give rise to the spread of the virus, and it was not the weekend that had to be declared but the @ * quarantine * quarantine @.'', \textbf{``In-favor quarantine''},
    \item ``Vacation would only give rise to the spread of the virus, and it was not the weekend that had to be declared but the @ * quarantine * quarantine @.'', \textbf{``Negative to quarantine''},
    \item ``Vacation would only give rise to the spread of the virus, and it was not the weekend that had to be declared but the @ * quarantine * quarantine @.'', \textbf{``Neutral to quarantine''},
    \item ``Vacation would only give rise to the spread of the virus, and it was not the weekend that had to be declared but the @ * quarantine * quarantine @.'', \textbf{``Positive to quarantine''}.
\end{itemize}

The stance or premise label was chosen as the one where the corresponding input example had the maximum softmax output.

\paragraph{sevastyanm (vyatsu)} This participant utilized pre-trained ruRoberta-large language model\footnote{\href{https://huggingface.co/sberbank-ai/ruRoberta-large}{https://huggingface.co/sberbank-ai/ruRoberta-large}} which was trained on additional data obtained from ``PersEssays\_Russian'' and ``ArgMicro\_Russian'' datasets \cite{Fishcheva-etal2021} with similar annotation schemes. Both datasets were united by argumentative discourse units and used to train model to solve 4-class classification problem. 

First, ruRoberta-large was fine-tuned on the united ``PersEssays\_Russian'' and ``ArgMicro\_Russian'' dataset with 8,780 units. Then, the model was fine-tuned separately on each of 6 tasks from the competition dataset \textit{('masks\_stance', 'masks\_argument', 'quarantine\_stance', 'quarantine\_argument', 'vaccines\_stance', 'vaccines\_argument')}. For final class prediction the participant used token averaging and 2-layer linear neural network classifier. All models trained with $learning\_rate = 10^{-5}$ and $weight\_decay = 0.01$.

For the model trained on the additional dataset the participant used the following hyperparameters: $input\_size=70, num\_epochs=3, batch\_size=16$. For model trained on the RuArg dataset hyperparameters were as follows:  $input\_size=100, batch\_size=32, num\_epochs=\{2, 2, 4, 4, 4, 7\}$ for \textit{'masks\_stance', 'masks\_argument', 'quarantine\_stance', 'quarantine\_argument', 'vaccines\_stance', 'vaccines\_argument'} respectively.

\paragraph{iamdenay (IICT)} This team used the pre-trained Crosslingual RoBERTa-large model and fine-tuned on the augmented data. They mostly augmented the data containing stances and arguments about \textit{``quarantine''}. For the augmentation the participants used mT5 model to paraphrase sentences in order to increase the size of the text set. To increase accuracy of the proposed method they used six different models, one per task \{\textit{'masks\_stance', 'masks\_argument', 'quarantine\_stance', 'quarantine\_argument', 'vaccines\_stance', 'vaccines\_argument'}\}.

\paragraph{ursdth} 
This team proposed a pipeline-based framework for the classification of texts with or without recognizable rhetorical structure. The first stage involved fine-tuning sequential model on the classification dataset including texts of different lengths and complexity. In the second stage, they frozen the base model and then trained a discourse-aware neural module on top of it for the classification of texts with discourse structure.

They used pre-trained Conversational RuBERT for the discourse unit classification. For texts with automatically recognizable discourse structure, they proposed a relation-aware Tree-LSTM over the discourse units' class predictions. Stance and premise labels were predicted jointly. 

Both development and test datasets were treated as unseen, and the official development dataset was not used for the parameters adjustment. The predictions were obtained by averaging outputs from five models trained on cross-validation during experiments over labeled data. This is similar to an ensemble, where each model is trained using 80\% of the train data.

\paragraph{sopilnyak (auteam)} 	This team started with training a classifier to detect irrelevant sentences for each sub-task. They applied binary Logistic Regression classifier trained on TF-IDF features, calculated from BPE tokens.

Then they excluded irrelevant sentences and further trained the models (for each sub-task separately) as a blend of:
\begin{enumerate}
    \item fine-tuned ruRoberta-large from Sber AI with a two-layer classification head on top. They unfrozen 30 top layers and used very low learning rate ($5\cdot10^{-6}$) to prevent model from over-fitting on a small dataset. Also they utilized weighted cross-entropy loss so that the results on unbalanced dataset would be more accurate.
    \item Logistic Regression classifier on TF-IDF features calculated on BPE tokens.
\end{enumerate}

\paragraph{kazzand} This participant applied Transformer-based deep text feature extraction and hierarchical classification. Firstly, they trained simple TF-IDF + Logistic Regression pipeline for each text type (masks, quarantine, vaccines). Secondly, they trained 6 separate models for each task \{\textit{'masks\_stance', 'masks\_argument', 'quarantine\_stance', 'quarantine\_argument', 'vaccines\_stance', 'vaccines\_argument'}\} using Sentence-BERT for embeddings computation served as input to the Logistic Regression or KNN model.

\paragraph{invincible} The first step for the team was a preprocessing: they removed punctuation symbols, converted text to lowercase, and removed special symbols including the “[USER]” substring. They further used the DistilRuBERT model\footnote{\href{https://huggingface.co/DeepPavlov/distilrubert-tiny-cased-conversational}{https://huggingface.co/DeepPavlov/distilrubert-tiny-cased-conversational}} to vectorize the text into a vector of numbers and saved as a row of the new matrix.
This feature matrix was used as an input for the classification models.

Overall, there were nine models, three for each topic \{``masks'', ``vaccines'', ``quarantine''\}. The initial data were separated into three subsets corresponding to each topic.
Then the following algorithm was applied: first, SVM model (with sigmoid kernel and balanced target) detected irrelevant sentences for each topic and classified them as ``irrelevant'' for both stance and argument types. Then for positive-classified sentences, two neural network models were applied. They consisted of Flatten layer, Dense layer with ReLU activation function, Dropout layer and final Dense layer with Sigmoid activation function. They used the following hyperparameters: $optimizer=$``$adam$'', $loss=``sparse\_categorical\_crossentropy$'', $num\_epochs=5$.

Importantly, since the classes are highly unbalanced (labels ``for'' and ``against'' are highly underrepresented), a random oversampling was applied to all dataframes before fitting the models. Accurate class-balancing allowed improving both scores significantly.

\section{Results and Discussion}
\label{sec:results}
\begin{table}[t]
\begin{center}
\resizebox{\textwidth}{!}{
\begin{tabular}{|c|c|r|r|r|r|r|r|}
\toprule 
\# &  Participant &  Base Transformer model &  \begin{tabular}[c]{@{}c@{}}Additional\\ data\end{tabular} &  Stance F1-score &  \# & Premise F1-score & \# \\ \midrule
1 & camalibi & covid-twitter-bert-v2 & Yes & 0.6968 & 1 & 0.7404 & 1 \\
2  & sevastyanm & RuRoBERTa-large & Yes  & 0.6815 & 2 & 0.7235 & 2 \\
3  & iamdenay   & RuRoBERTa-large & Yes  & 0.6676 &3 & 0.6555 &4 \\
4  & ursdth    & RuBERT Conversational & No   & 0.6573 & 4 &  0.7064 &3\\
5  & sopilnyak  & RuRoBERTa-large & No  & 0.5603 &5 & 0.4338 &10 \\
6  & kazzand   & Sentence-BERT & No  & 0.5552 &6 & 0.5603 &6\\
7  & morty     & n/a & n/a  &  0.5353 & 7 & 0.5453 &7\\
8  & invincible  & RuBERT Conversational & Yes & 0.5286 &8 & 0.5428 &8\\
9   & dr         & n/a &  n/a  & 0.4750 &9 & 0.6036 &5 \\ \midrule
10 & baseline    &  ruBERT &  No & 0.4180 &10 &  0.4355 &9 \\
\bottomrule
\end{tabular}}
\end{center}
\caption{Competition results of the participant systems. \linebreak The places of participants for each sub-task are indicated in the brackets.}
\label{tab:stance}
\end{table}

\begin{figure}[t]
  \centering
  \includegraphics[width=\linewidth]{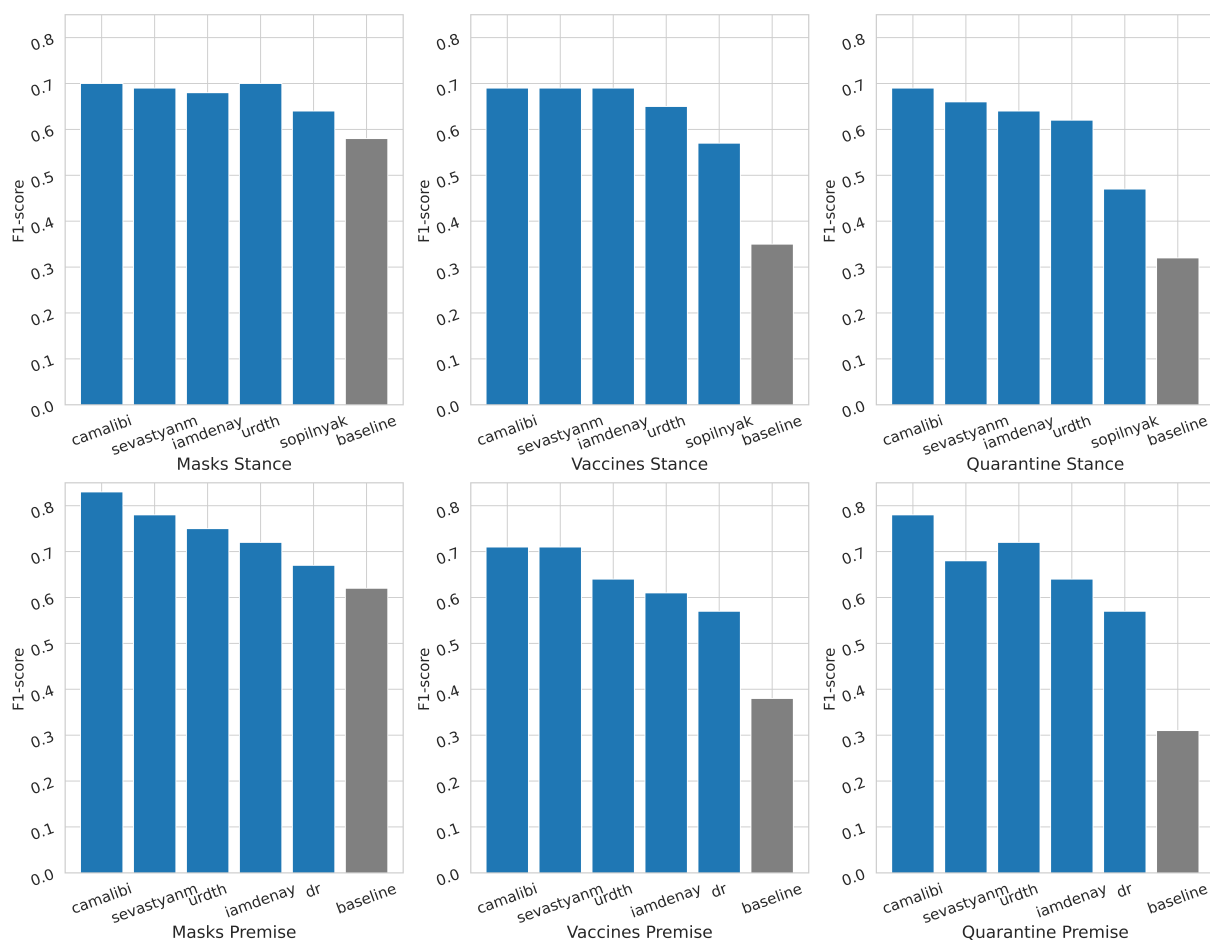}
  \caption{Results for the first top 5 participants for each sub-task.}
  \label{fig:results}
\end{figure}

Table \ref{tab:stance} presents respectively the results for ``stance detection'' and ``premise classificaiton'' tracks. 

All the results are quite stable for both sub-tasks, only \textbf{sopilnyak} did not manage to overcome the premise baseline, demonstrating high results (top-5) at the stance detection sub-task. The range of the models used to solve the task is not wide: the participants choose between (ru)BERT, (crosslingual)RoBERTa(-large) and old good Logistic Regression model.

In comparison to the baseline, all the participants trained classification models separately for each sub-task. Evidently, multitask classification is more challenging than training classification models separately.

Interestingly, several best results were obtained with the help of the additional datasets or/and data augmentation (\textbf{camalibi} -- top-1, \textbf{sevastyanm} -- top-2, \textbf{iamdenay} -- top-3 for stance detection and top-4 for premise classification, and also \textbf{invincible} -- top-8). Top-1 \textbf{camalibi} used the special version of BERT model in which domain-oriented dataset was actually integrated; top-2 \textbf{sevastyanm} utilized additional dataset, top-3/top-4 \textbf{iamdenay} applied mT5 for paraphrase generation, top-8 \textbf{invincible} used random oversampling. From these observations we can assume that any kind of additional data is beneficial for these tasks, however, the more diverse the data is, the better.

The most different and outstanding approach in comparison to other participants was presented by the winner system of \textbf{camalibi}. This participant applied NLI method which performed best for both sub-tasks. Moreover, model trained on the English language was applied, therefore, \textbf{camalibi} did translate the whole dataset for the task.

We also compared the detailed results for the top 5 systems and the baseline. The scores are presented in Appendix \ref{appendix}. From Figure \ref{fig:results} we can see that F1-scores for Premise Classification are slightly higher than for Stance Detection. The task of Stance Detection is equally hard for all three topics, whereas we can see that the scores for Masks Premise and Quarantine Premise are higher than Vaccines Premise results. It can be seen that the difference between the top 3 participants are not very much different from each other. As for the baseline results, we can see that the results for vaccine and quarantine are two times lower than the results of the (at least) top 4 participants. At the same time, Masks Stance and Premise results are higher than for vaccines and quarantine and not significantly different from the top results. To sum up, we can conclude that the algorithms for the top 3 results demonstrate similar results across different subsets.

\section{Conclusion}
\label{sec:conclusion}

We present the results of the first shared task on Argument Mining for Russian.
For this shared task, we created a new dataset on the vital COVID-19 topic. We introduce and rely on the following claims: ``Vaccination is beneficial for society'', ``The introduction and observance of quarantine is beneficial for society'', and ``Wearing masks is beneficial for society''.

Overall, 13 teams participated in the shared task, and more than half of them outperformed the baseline model. The winning system in both sub-tasks used the NLI (Natural Language Inference) variant of the BERT architecture, automatic translation into English to apply a specialized BERT model, pretrained on Twitter posts discussing COVID-19, and additional masking of target entities. This system showed  for stance detection F1-score of 0.6968, for premise extraction  F1-score of 0.7404 which considerably outperforms the proposed BERT-based baseline (F1-scores of 0.4180 and 0.4355, respectively).

According to the provided results, we see that the argument mining is a feasible task, especially on the COVID-19 dataset. All the data and codes are available online.\footnote{\href{https://github.com/dialogue-evaluation/RuArg}{https://github.com/dialogue-evaluation/RuArg}} We hope that these materials will help to foster further research and developments in the area of argument mining for the Russian language.

As future work, we see it promising to explore more complex argument mining setups such as sequence tagging \cite{chernodub-etal-2019-targer} or information retrieval \cite{Bondarenko-etal:2020}.

\section*{Acknowledgements}
The work of Natalia Loukachevitch in selection of users' comments and stance annotation is supported by Russian Foundation for Basic Research (project N 20-04-60296).
The work of Evgeny Kotelnikov on premise annotation is supported by Russian Science Foundation (project N 22-21-00885 \footnote{\href{https://rscf.ru/en/project/22-21-00885/}{https://rscf.ru/en/project/22-21-00885/}}).

\bibliography{dialogue}
\bibliographystyle{dialogue}


\clearpage
\onecolumn
\phantomsection
\appendix
\section{All results}\label{appendix}

From the detailed results for each label we can see that the top 1 result is not always the best approach: many models from the top list perform equally well or even outperform the winner in many cases. For instance, from Tables \ref{m_s}, \ref{v_s} and \ref{v_p} we can see that \textbf{sevastyanm} and \textbf{urdth} are very competitive approaches. The top 1 ranking is achieved by demonstrating stable results across different subsets and good (normally best) precision scores. Interestingly, the best recall scores on quarantine subsets and the vaccines premise subset is achieved by the baseline (more than 0.95 points).

\begin{table}[h!]
\resizebox{\textwidth}{!}{
\begin{tabular}{|l|l|l|l|l|l|l|l|l|l|l|l|l|l|l|l|}
\toprule
\multirow{2}{*}{Method} & \multicolumn{3}{c|}{Irrelevant} & \multicolumn{3}{c|}{Against} & \multicolumn{3}{c|}{Other} & \multicolumn{3}{c|}{For} & \multicolumn{3}{c|}{Macro Average} \\ \cmidrule{2-16}
                        & Pr     & R     & F1    & Pr    & R     & F1    & Pr    & R     & F1    & Pr    & R     & F1    & Pr       & R        & F1      \\
                        \midrule
camalibi                & \textbf{1.00}   & \textbf{1.00}  & \textbf{1.00} &  \textbf{0.68} & \textbf{0.66} & \textbf{0.67} & 0.80  & 0.83 &  \textbf{0.81} &  \textbf{0.65} &  0.61 &  0.63    &  \textbf{0.71}  & \textbf{0.70}  & \textbf{0.70}    \\
sevastyanm              & \textbf{1.00}   & \textbf{1.00}  & \textbf{1.00}  & 0.65 & 0.63  & 0.64  &  0.80 & 0.82 & \textbf{0.81} & 0.62  & 0.62  & 0.62  &   0.69   &   0.69  &    0.69  \\
iamdenay                & \textbf{1.00}   & 0.99  & \textbf{1.00}  & 0.67  & 0.65  &  0.66 & 0.77 & 0.81 & 0.79  & 0.61 &  0.56 & 0.58 &  0.68   &  0.67   & 0.68    \\
urdth                   & \textbf{1.00}   & \textbf{1.00}  & \textbf{1.00}  & 0.62 & 0.65  & 0.63 & \textbf{0.82}  & 0.80  & \textbf{0.81}  &  0.64 & \textbf{0.65}  &  \textbf{0.65} &   0.69   &  \textbf{0.70}   &   \textbf{0.70}  \\
sopilnyak                   & 0.98   & \textbf{1.00}  & 0.99  & 0.64 & 0.50  & 0.56 & 0.77  & \textbf{0.86}  & \textbf{0.81}  &  0.63 & 0.50  &  0.56 &   0.68   &  0.62   &   0.64  \\
\midrule
baseline                & 0.99   & 0.99  & 0.99  & 0.46  & 0.49  &  0.47 & 0.77 & 0.77 &  0.77 &  0.50 &  0.48 &  0.49 &  0.58  &   0.58  & 0.58  \\
\bottomrule
\end{tabular}}
\caption{Results for masks stance for top-5 participants.}
\label{m_s}
\end{table}

\begin{table}[h!]
\resizebox{\textwidth}{!}{
\begin{tabular}{|l|l|l|l|l|l|l|l|l|l|l|l|l|l|l|l|}
\toprule
\multirow{2}{*}{Method} & \multicolumn{3}{c|}{Irrelevant} & \multicolumn{3}{c|}{Against} & \multicolumn{3}{c|}{Other} & \multicolumn{3}{c|}{For} & \multicolumn{3}{c|}{Macro Average} \\ \cmidrule{2-16}
                        & Pr     & R     & F1    & Pr    & R     & F1    & Pr    & R     & F1    & Pr    & R     & F1    & Pr       & R        & F1      \\
                        \midrule
camalibi                & \textbf{1.00}   & \textbf{1.00}  & \textbf{1.00}  & 0.78  & \textbf{0.73}  & \textbf{0.76}  & \textbf{0.94}  & 0.94  & \textbf{0.94}  & \textbf{0.73}  & \textbf{0.84}  & \textbf{0.78}  & \textbf{0.82}     & \textbf{0.84}     & \textbf{0.83}    \\
sevastyanm              & \textbf{1.00}   & \textbf{1.00}  & \textbf{1.00}  & \textbf{0.85}  & 0.61  & 0.71  & 0.93  & \textbf{0.96}  & \textbf{0.94}  & 0.65  & 0.70  & 0.67  & 0.81     & 0.76     & 0.78    \\
urdth                   & \textbf{1.00}   & \textbf{1.00}  & \textbf{1.00}  & 0.67  & \textbf{0.73}  & 0.70  & 0.93  & 0.93  & 0.93  & 0.68  & 0.60  & 0.64  & 0.76     & 0.75     & 0.75    \\
iamdenay                & \textbf{1.00}   & \textbf{1.00}  & \textbf{1.00}  & \textbf{0.85}  & 0.47  & 0.61  & 0.89  & \textbf{0.96}  & 0.93  & 0.64  & 0.62  & 0.63  & 0.79     & 0.68     & 0.72    \\
dr                & \textbf{1.00}   & 0.99  & \textbf{1.00}  & 0.77  & 0.47  & 0.58  & 0.88  & \textbf{0.96}  & 0.92  & 0.59  & 0.46  & 0.52  & 0.75     & 0.63     & 0.67    \\
\midrule
baseline                & \textbf{1.00}   & 0.99  & 0.99  & 0.56  & 0.41  & 0.48  & 0.89  & 0.92  & 0.91  & 0.44  & 0.51  & 0.47  & 0.63     & 0.61     & 0.62  \\
\bottomrule
\end{tabular}}
\caption{Results for masks premise for top-5 participants.}
\label{m_p}
\end{table}

\begin{table}[h!]
\resizebox{\textwidth}{!}{
\begin{tabular}{|l|l|l|l|l|l|l|l|l|l|l|l|l|l|l|l|}
\toprule
\multirow{2}{*}{Method} & \multicolumn{3}{c|}{Irrelevant} & \multicolumn{3}{c|}{Against} & \multicolumn{3}{c|}{Other} & \multicolumn{3}{c|}{For} & \multicolumn{3}{c|}{Macro Average} \\ \cmidrule{2-16}
                        & Pr     & R     & F1    & Pr    & R     & F1    & Pr    & R     & F1    & Pr    & R     & F1    & Pr       & R        & F1      \\
                        \midrule
camalibi                & \textbf{1.00}   & \textbf{1.00}  & \textbf{1.00}  & 0.78  & 0.58 & 0.67 & 0.72 &  \textbf{0.86} & \textbf{0.79} & \textbf{0.71}  & 0.56  & \textbf{0.63}  &  \textbf{0.74}    &     0.67 &   \textbf{0.69}  \\
sevastyanm              & \textbf{1.00}   & \textbf{1.00}  & \textbf{1.00}  & 0.71 & \textbf{0.74}  & \textbf{0.73}  &  \textbf{0.75} & 0.75 & 0.75 & 0.63  & \textbf{0.59}  & 0.61  &   0.70   &  \textbf{0.69}   & \textbf{0.69}     \\
iamdenay                & \textbf{1.00}   & \textbf{1.00}  & \textbf{1.00}  & \textbf{0.79} & 0.60  & 0.69  & 0.71 & \textbf{0.86}  & 0.78  & 0.70 &  0.52 & 0.60 &  0.73   &    0.66 &  \textbf{0.69}   \\
urdth                   & \textbf{1.00}   & \textbf{1.00}  & \textbf{1.00}  & 0.67 & 0.57  & 0.61 &  0.70 &  0.82 & 0.76  & 0.68  & 0.52  &  0.59 &   0.68   &   0.64  &  0.65   \\
sopilnyak                   & 0.99   & \textbf{1.00 } & \textbf{1.00}  & 0.61 & 0.43  & 0.51 &  0.67 &  0.79 & 0.72  & 0.54  & 0.41  &  0.47 &   0.61   &   0.55  &  0.57   \\
\midrule
baseline                & 0.99   & \textbf{1.00}  & 0.99  & 0.43  & 0.15  &   0.22 & 0.56 & 0.85 & 0.67  &  0.38 &  0.11 & 0.17  &  0.46  &  0.37   & 0.35  \\
\bottomrule
\end{tabular}}
\caption{Results for vaccines stance for top-5 participants.}
\label{v_s}
\end{table}

\begin{table}[h!]
\resizebox{\textwidth}{!}{
\begin{tabular}{|l|l|l|l|l|l|l|l|l|l|l|l|l|l|l|l|}
\toprule
\multirow{2}{*}{Method} & \multicolumn{3}{c|}{Irrelevant} & \multicolumn{3}{c|}{Against} & \multicolumn{3}{c|}{Other} & \multicolumn{3}{c|}{For} & \multicolumn{3}{c|}{Macro Average} \\ \cmidrule{2-16}
                        & Pr     & R     & F1    & Pr    & R     & F1    & Pr    & R     & F1    & Pr    & R     & F1    & Pr       & R        & F1      \\
                        \midrule
camalibi                & \textbf{1.00}   & \textbf{1.00}  & \textbf{1.00}  &  \textbf{0.79}  & 0.57 & \textbf{0.67}  & \textbf{0.89} & 0.94  &  \textbf{0.92} & 0.55  &  0.52 & 0.54  &  \textbf{0.75}    &   0.68   &   \textbf{0.71}  \\
sevastyanm              & \textbf{1.00}   & \textbf{1.00}  & \textbf{1.00}  & 0.63 & \textbf{0.59}  & 0.61  &  0.90 & 0.92  & 0.91 & \textbf{0.67}  &  \textbf{0.57} &  \textbf{0.62} &   0.73   &    \textbf{0.69} &   \textbf{0.71} \\
urdth                   & \textbf{1.00}   & \textbf{1.00}  & \textbf{1.00}  & 0.56 & \textbf{0.59}  & 0.58 &  \textbf{0.89} &  0.88 &  0.88 & 0.45  &  0.48 & 0.47 &    0.64  &   0.65  &   0.64  \\
iamdenay                & \textbf{1.00}   & \textbf{1.00}  & \textbf{1.00}  & 0.60 &  0.50  &  0.55  & 0.86 & 0.85  &  0.86 & 0.34 &  0.52 &  0.42 &  0.60   & 0.63    &  0.61   \\
dr                & \textbf{1.00}   & \textbf{1.00}  & \textbf{1.00}  & 0.55 &  0.39  &  0.46  & 0.85 & 0.91  &  0.88 & 0.39 &  0.33 &  0.36 &  0.60   & 0.54    &  0.57   \\
\midrule
baseline                & \textbf{1.00}   & 0.99  & 0.99  &  0.43 &  0.11 &  0.18 & 0.80 & \textbf{0.95} &  0.87 &  0.33 &  0.05 & 0.08  &  0.52  &  0.37   &  0.38 \\
\bottomrule
\end{tabular}}
\caption{Results for vaccines premise for top-5 participants.}
\label{v_p}
\end{table}

\begin{table}[h!]
\resizebox{\textwidth}{!}{
\begin{tabular}{|l|l|l|l|l|l|l|l|l|l|l|l|l|l|l|l|}
\toprule
\multirow{2}{*}{Method} & \multicolumn{3}{c|}{Irrelevant} & \multicolumn{3}{c|}{Against} & \multicolumn{3}{c|}{Other} & \multicolumn{3}{c|}{For} & \multicolumn{3}{c|}{Macro Average} \\ \cmidrule{2-16}
                        & Pr     & R     & F1    & Pr    & R     & F1    & Pr    & R     & F1    & Pr    & R     & F1    & Pr       & R        & F1      \\
                        \midrule
camalibi                & 0.99   & \textbf{1.00}  & \textbf{1.00}  &   \textbf{0.88} &  0.35 & \textbf{0.50} & 0.84 & 0.83  & 0.83 & \textbf{0.70}  & 0.81  &  \textbf{0.75} &   \textbf{0.80}   &  \textbf{0.66}    &  \textbf{0.69}   \\
sevastyanm              & 0.99   & \textbf{1.00}  & \textbf{1.00}  & 0.57 &  \textbf{0.40} &  0.47 & \textbf{0.85}  & 0.78  & 0.82 &  0.62 & 0.79  &  0.70 &  0.68    &   \textbf{0.66}  &   0.66  \\
iamdenay                & 0.99   & \textbf{1.00}  & 0.99  & 0.67 &  0.25 & 0.36  & 0.84 & 0.85 &  \textbf{0.84} & \textbf{0.70} &  0.75 & 0.72 &  0.73   &  0.62   &   0.64  \\
urdth                   & 0.99   & \textbf{1.00}  & \textbf{1.00}  & 0.56 & 0.35  & 0.43 & 0.84  & 0.71  &  0.77 &  0.56 &  \textbf{0.82} & 0.67  &   0.65   &   0.63  &  0.62   \\
sopilnyak                   & 0.98   & \textbf{1.00}  & 0.99  & 0.00 & 0.00  & 0.00 & 0.77  & 0.80  &  0.78 &  0.59 &  0.67 & 0.63  &   0.45   &   0.49  &  0.47   \\
\midrule
baseline                & \textbf{1.00}   & 0.99  & \textbf{1.00} & 0.00  &  0.00 &  0.00 & 0.65 & \textbf{0.97} & 0.77  &  0.62 &  0.11 &  0.19 &  0.42  &   0.36  & 0.32  \\
\bottomrule
\end{tabular}}
\caption{Results for quarantine stance for top-5 participants.}
\label{q_s}
\end{table}

\begin{table}[h!]
\resizebox{\textwidth}{!}{
\begin{tabular}{|l|l|l|l|l|l|l|l|l|l|l|l|l|l|l|l|}
\toprule
\multirow{2}{*}{Method} & \multicolumn{3}{c|}{Irrelevant} & \multicolumn{3}{c|}{Against} & \multicolumn{3}{c|}{Other} & \multicolumn{3}{c|}{For} & \multicolumn{3}{c|}{Macro Average} \\ \cmidrule{2-16}
                        & Pr     & R     & F1    & Pr    & R     & F1    & Pr    & R     & F1    & Pr    & R     & F1    & Pr       & R        & F1      \\
                        \midrule
camalibi                & 0.99   & \textbf{1.00}  & \textbf{1.00}  & \textbf{0.60}  & 0.27 & 0.37 & 0.91 & 0.96  & \textbf{0.93} &  \textbf{0.85} & 0.68  & 0.76  &  \textbf{0.84}    & 0.73     &    \textbf{0.77} \\
sevastyanm              & 0.99   & 0.99  & 0.99  & 0.44 & \textbf{0.50}  & \textbf{0.47}  & 0.91  & 0.92 &  0.92 & 0.72  & 0.62  & 0.67  &   0.69   &    0.68 &   0.68  \\
urdth                   & 0.99   & \textbf{1.00}  & \textbf{1.00}  & 0.47  &  0.41 & 0.44 &  \textbf{0.95} &  0.92 & \textbf{0.93}  &  0.73 &  \textbf{0.88} & \textbf{0.80}  &    0.72  &  \textbf{0.74}  &   0.72  \\
iamdenay                & 0.99   & \textbf{1.00}  & \textbf{1.00}  & 0.33  & 0.41  &  0.37 & 0.91 & 0.90 & 0.91  & 0.70 & 0.60  & 0.65 &    0.65 &  0.64   &  0.64   \\
dr                & 0.99   & \textbf{1.00}  & \textbf{1.00}  & 0.42  & 0.23  &  0.29 & 0.88 & 0.94 & 0.91  & 0.66 & 0.42  & 0.51 &    0.65 &  0.53   &  0.57  \\
\midrule
baseline                & \textbf{1.00}   & 0.99  & \textbf{1.00}  & 0.00 & 0.00  & 0.00  & 0.82 & \textbf{0.99}  &  0.90 &  0.50 &  0.02 &  0.04 &  0.44  &    0.34 & 0.31  \\
\bottomrule
\end{tabular}}
\caption{Results for quarantine premise for top-5 participants.}
\label{q_p}
\end{table}

\end{document}